\documentclass[letterpaper, 10 pt, conference]{ieeeconf}  %

\IEEEoverridecommandlockouts                              %

\usepackage{xcolor}         %
\usepackage{symbols}
\usepackage{subcaption}
\usepackage{lipsum}
\usepackage{graphicx}
\usepackage{amsmath}
\usepackage{amssymb}
\usepackage{dsfont}

\usepackage{multirow}
\usepackage{longtable}
\usepackage{booktabs}
\usepackage{hyperref}
\usepackage{amssymb}
\usepackage{amsfonts}
\usepackage{epigraph}
\usepackage{bbm}
\usepackage{algorithm}
\usepackage{algpseudocode}
\usepackage{amssymb}
\usepackage{tabularx}

\title{\LARGE \bf
Exploitation-Guided Exploration for Semantic Embodied Navigation
}

\author{Justin Wasserman$^{1,2}$ Girish Chowdhary$^{1}$ Abhinav Gupta$^{2}$ Unnat Jain$^{2,3}$%
\thanks{$^{1}$ University of Illinois at Urbana-Champaign
}%
\thanks{$^{2}$ Carnegie Mellon University\quad $^{3}$ FAIR at Meta
}%
}

\DeclareMathOperator{\E}{\mathbb{E}}
\begin{document}

\maketitle
\thispagestyle{empty}
\pagestyle{empty}

\begin{abstract}
In the recent progress in embodied navigation and sim-to-robot transfer, modular policies have emerged as a de facto framework. However, there is more to compositionality beyond the decomposition of the learning load into modular components. In this work, we investigate a principled way to syntactically combine these components. Particularly, we propose Exploitation-Guided Exploration (\ourmethodshort) where separate modules for exploration and exploitation come together in a novel and intuitive manner. We configure the exploitation module to take over in the deterministic final steps of navigation \ie when the goal becomes visible. Crucially, an exploitation module teacher-forces the exploration module and continues driving an overridden policy optimization. 
\ourmethodshort, with effective decomposition and novel guidance, improves the state-of-the-art performance on the challenging object navigation task from 70\% to 73\%.
Along with better accuracy, through targeted analysis, we show that \ourmethodshort is also more efficient at goal-conditioned exploration. 
Finally, we show sim-to-real transfer to robot hardware and \ourmethodshort performs over two-fold better than the best baseline from simulation benchmarking.\\
Project page: \href{https://xgxvisnav.github.io/}{xgxvisnav.github.io}

\end{abstract}

\section{Introduction}
\epigraph{The meaning of a whole is a function of the meanings of the parts and of \textbf{the way they are syntactically combined}.}{\textit{Principle of Compositionality (Frege's Principle)}}

Consider the `planning' problem you underwent when finalizing your last vacation. There is usually plenty of uncertainty in the decision-making -- where to go, is the weather good, managing the budget, booking ground transport, \etc. We tackle this uncertainty by breaking the problem down into sub-parts and then offload some of these sub-parts to specialists or expert websites.
However, there is an `additional order' that goes beyond offloading sub-parts to these reliable and modular solutions. The very fact that you know these modular solutions exist helps you make more ambitious plans, be more creative, and likely travel more often and better than without this information. Generally speaking, division of work or \textit{modularity is only one aspect of a compositional approach}. The very awareness of our repertoire of skills or (associated modules) helps manage resources to plan uncertain aspects of planning better. This is true about compositional structures and software packages allowing us to think more creatively in tackling research as well. This intuition of our day-to-day intelligence guides our alternate take on visual robot navigation or \textit{embodied navigation} which we discuss in this work.

While compositionality has been thoroughly investigated with classical~\cite{fodor1988connectionism} and neural lenses~\cite{andreas2016neural,AndreasCVPR2016} in natural language processing, the utility of modular structure in learning-based navigation has only recently been caught attention, with the development of reproducible task definitions and platforms~\cite{ai2thor,AllenAct,chang2017matterport3d,habitat19iccv,batra2020objectnav,hahn2021no,Szot2023AdaptiveCI,deitke2022retrospectives}. Particularly, the decomposition of exploration and exploitation policies (frequently dubbed as global and local policies~\cite{Chaplot2020Explore,chaplot2020object}) is a common and intuitive one. Also, the exploitative final steps after exploration (also dubbed as last-mile navigation~\cite{ye2021auxiliary,chattopadhyay2021robustnav,wasserman2022lastmile}) are shown to be better tackled by principled geometric vision or homography-based visuo-motor servoing (see~\cite{wasserman2022lastmile} for a related study). 
As motivated above, we believe there is more to be tapped in compositionality beyond policy decomposition. To this end, we investigate the research question: \textit{``Can modular navigation agents learn better exploration when guided by their exploitation abilities?''}

\begin{figure*}[t]
    \centering
    \vspace{1mm}
    \includegraphics[width=0.99\textwidth]{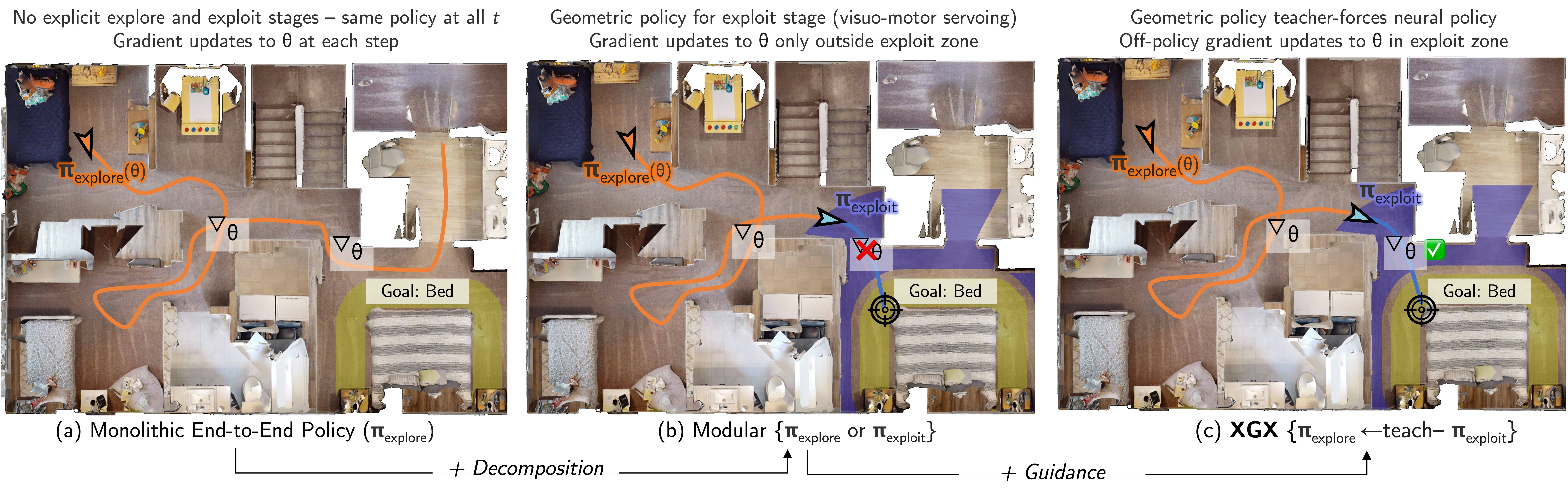}

    \caption{
    \textbf{Overview of \ourmethod (\ourmethodshort).} Within learning-based visual navigation (\eg ``navigate to category: \textit{Bed}''), (a) most prior works adopt a neural policy, trained end-to-end. (b) Some works employ a modular policy, with a dedicated module to explore the environment and another for local navigation near the goal; we devise a simple and effective decomposition in \ourmethodshort. (c) Unlike prior work, \ourmethodshort enables the exploitation module to guide exploration module via off-policy updates and teacher forcing. Both pillars of \ourmethodshort (\textit{policy decomposition} and \textit{guidance}) are detailed in \secref{sec:approach}.
    }
    \vspace{-5mm}
    \label{fig:lmn_td}
\end{figure*}

Given that we as a research community have already built most of the blocks, the investigation of this research question is rather intuitive and straightforward! For this study, we focus on the semantic visual navigation task of object-goal navigation~\cite{batra2020objectnav} where the agent’s objective is to navigate to a goal category (akin to ``find a chair''), which is a standardized and reproducible benchmark~\cite{habitat-challenge-2021}. 
Our policy \textit{decomposition} employs a state-of-the-art neural policy~\cite{ramrakhya2023pirlnav} for the exploration module. For the exploitation module, we devise a simple-and-effective geometric policy for visuo-motor servoing  the exploitation phase (steps after the goal is in view). 
Contrary to prior works, in \ourmethodshort, the exploitation module provides \textit{guidance} to the exploration module (implemented as teacher-forced~\cite{williams1989learning} variation of the Proximal Policy Optimization). Note, prior works tackling vision-based navigation via modular policy \textit{independently} optimizes modules, with no guidance or interplay in loss objectives.
\figref{fig:lmn_td} illustrates these two pillars of \ourmethodshort -- \textit{decomposition}, and \textit{guidance}.

We quantify the utility of \ourmethodshort with benchmarking in simulation and sim-to-real transfer. Summary of our contributions: (1) on the competitive object-goal navigation (HM3D) benchmark, across baselines from several learning paradigms and frontier-based methods, \ourmethodshort leads to a 70\% $\rightarrow$ 73\% increase in success rate over the previous state-of-the-art ; (2) \ourmethodshort needs lesser surveying while increasing success (captured via a new metric that balances task success with exploration efficiency), (3) rigorous real-robot experiments showing over two-fold improvement over best-performing baselines from simulation; (4) Detailed ablations, quantitative analysis of efficiency, and error modes.

\section{Related Work}
\label{sec:related_works}

\noindent\textbf{Embodied Navigation.} 
A popular paradigm for solving object-goal navigation has been to learn directly from simulation. This has been achieved through modular~\cite{hahn2021no,wani2020multion} approaches, end-to-end~\cite{wijmans2019dd} approaches, and by fine-tuning to the environment~\cite{ramrakhya2023pirlnav}. DDPPO~\cite{wijmans2019dd} has been used as a baseline for many experiments for its success in solving point-goal navigation. They train a policy via a decentralized and distributed update rule, allowing them to train a policy in the high-fidelity simulator, AIHabitat~\cite{habitat19iccv,szot2021habitat,puig2023habitat}, with over 2.5 billion frames. Habitat-Web~\cite{ramrakhya2022habitat} was able to achieve competitive performance on the 2022 Habitat Challenge by utilizing imitation learning over trajectories collected from a human. In contrast to directly using the simulator for fully training a policy from scratch, several methods have also attempted to use pretraining~\cite{yadav2022OVRL,yadav2023ovrl,khandelwal2022simple} to improve the navigation results. OVRL-v2 has achieved the current SOTA performance on the image-goal and object-goal by first pretraining a transformer head via masked autoencoding on HM3D~\cite{ramakrishnan2021habitat} and Gibson~\cite{xia2018gibson}. Moving further away from simulation, a number of methods have been proposed to solve visual navigation tasks without any training in simulation. These zero-shot methods include frontier-based exploration~\cite{gervet2022navigating} or heuristics~\cite{wasserman2022lastmile} in conjunction with a module to solve the navigation task, as well as learning from real-world demonstrations~\cite{hahn2021no,chang2020semantic}.

\noindent\textbf{Modular Policies in Visual Navigation.}
Using modular policies is a popular paradigm for solving the visual-navigation task. Previous methods such as NRNS~\cite{hahn2021no} use a modular strategy to build a topological map and with a Graph Neural Network predicts where on the topological map the agent should explore to find the image-goal. They propose a neural exploitation policy to complete the task. In Chaplot~\etal~\cite{chaplot2020object} the authors optimize for exploration by training a semantic-map conditioned policy to predict a long-term goal of where in the map the agent should explore to. Once their agent is near the goal, they utilize a deterministic local policy to arrive at the goal. In NTS~\cite{chaplotNeuralTopologicalSLAM2020} the authors follow a similar paradigm by training a topological map to explore the environment and then using a local policy to reach a relative waypoint from the agent. Finally, in PONI~\cite{ramakrishnan2022poni}, the authors propose a potential function network that predicts ``where to look'' in the environment by predicting the most promising area along the boundaries of a map. The authors also use an analytical planner to predict an action to solve how to navigate to a given boundary. In all of these previous works, the authors keep the exploration and exploitation modules separated.
In contrast to these previous works, \pipeshort does not learn to explore separately from the exploitation module. Instead, we learn to explore the environment with feedback from our exploitation module included during training.

\noindent\textbf{Goal-Conditioned Navigation on Physical Robots.}
Several recent works study robot navigation and locomotion, particularly in sim-to-real settings~\cite{truong2023rethinking,gervet2022navigating,agarwal2022legged,fu2021coupling}.
This is an especially challenging task as real-world robots are susceptible to error modes not seen in simulation such as actuation and sensing noise, no access to ground truth data, and collecting trajectory examples is much slower~\cite{truong2021bi,chebotar2019closing,tobin2017domain}. Related works have studied image-goal and point-goal tasks and have shown transfer to a physical robot. ViKiNG~\cite{shah2022viking,shah2023vint} was proposed to solve the image-goal task at a kilometer scale by utilizing geographical hints through a top-down map. SLING~\cite{wasserman2022lastmile} was able to improve over all previous baselines in simulation and real by utilizing a geometric-based solution for solving last-mile navigation. TGSM~\cite{kim2023topological} utilizes a cross-graph mixer over a topological map that incorporates both image and object nodes to achieve SOTA performance on the image-goal task when a panoramic camera is utilized. However, in these tasks, an exact image or position of the goal is required. This does not allow for generalizing to our semantic visual navigation setting of navigating to a label such as ``chair''. %

\section{\pipe}
\label{sec:approach}
As visualized in~\figref{fig:lmn_td}, the key components of \ourmethod methodology are exploration module (\piex), phase transition to exploitation module (\pig), and optimization of the exploration module via guidance from the exploitation module. Further details regarding implementation and parameter choice are given in Appendix~\ref{sec:app-impl}.

\subsection{Exploration Module \texorpdfstring{(\piex)}{TEXT}}
\label{subsec:explore}
The neural architecture choice for the exploration module \piex is orthogonal to our novelty of \ourmethodshort (decomposition and guidance). To this end, we adopt the best-performing publicly available design -- PIRLNav by Ramrakhya~\etal~\cite{ramrakhya2023pirlnav}.
The architecture includes a ResNet~\cite{he2016deep} \ie convolutional blocks for visual encoding and gated-recurrent unit~\cite{ChoARXIV2015} for connecting observation across time. A multi-layer perceptron policy head on top outputs a categorical action distribution. For further details, kindly refer to~\cite{ramrakhya2023pirlnav}.
Taking lessons from their extensive benchmarking, 
we too warm-start the neural exploration module by imitation learning over offline demonstrations (from~\cite{ramrakhya2022habitat}) and then fine-tune with reinforcement learning via \ourmethodshort.
The decomposed policy transitions from the exploration module to the exploitation module, if certain criterion is met, which we explain next.

\subsection{Phase Transition} 
\label{subsec:phase}
An ideal transition from exploration to exploitation phases would be characterized by handing off deterministic final stage of navigation to the exploitation module. One way to realize this is by training the exploration module with an \textit{updated} reward signal, $\mathds{1} \text{\{Exploitation Module Succeeds\}}$ rather than direct success rewards akin to $\mathds{1} \text{\{Within } 1m \text{ of the goal\}}$.
However, calculating such an \textit{updated} reward signal, at every step, is computationally intractable (requires a full trajectory rollout at each step). We implement a tractable and effective approximation of this \textit{updated} reward signal. Intuitively, we check if the agent can `see' the goal and is close to it. If so, phase transition from exploration module to exploitation module should triggered.

Specifically, if semantic segmentation (of the agent's visual observation) contains the goal category and the agent is within a prescribed distance $\delta$ from the goal (as inferred from depth observation), the control transitions to the exploitation module. The set of states for which this transition will occur can be more mathematically described by the set: 
\begin{align}
    \mathcal{S}^{j}_{\text{exploit}} \triangleq \{ (s) \in \mathcal{S} : \text{dist}_j(s) \leq \delta,\;\; O_{j} \in f_{\text{semantic}}(s) \},
    \label{eq:set}
\end{align}
where $j$ is an index corresponding to the object-goal category labels $\{ O_1,\ldots, O_n\}$, and $f_{\text{semantic}}$ an off-the-shelf semantic segmentation model. In this set, which we call the `exploitation states', the agent employs simple and effective visuo-motor servoing \ie our exploitation module, detailed next.

\subsection{Exploitation Module \texorpdfstring{(\pig)}{TEXT}}
\label{subsec:exploit}
Once the goal is in sight, \ie, within exploitation state, visuo-motor servoing is both simple and effective. To this end, our exploitation module \pig can triangulate the goal, navigating to it, and executing the `stop' action. 
The exploitation module, utilizing the same off-the-shelf semantic segmentation model as phase transition, it transforms the RGB to a semantic mask of the goal object. Next, it lifts the 2D semantic mask to 3D by utilizing the depth mask. From this depth mask, and with knowledge of the camera intrinsics, a waypoint is calculated.
To navigate to this waypoint, direct planning works well. For this, we utilize a local metric map (free from the depth sensor) and employ a fast-marching method~\cite{Chaplot2020Explore,hahn2021no} for collision avoidance. 
Next, we include \ourmethodshort loss function and guidance-based policy updates. 
\begin{figure*}[h]
    \centering
    \vspace{2mm}
    \begin{subfigure}[b]{0.15\textwidth}
        \includegraphics[width=\linewidth]{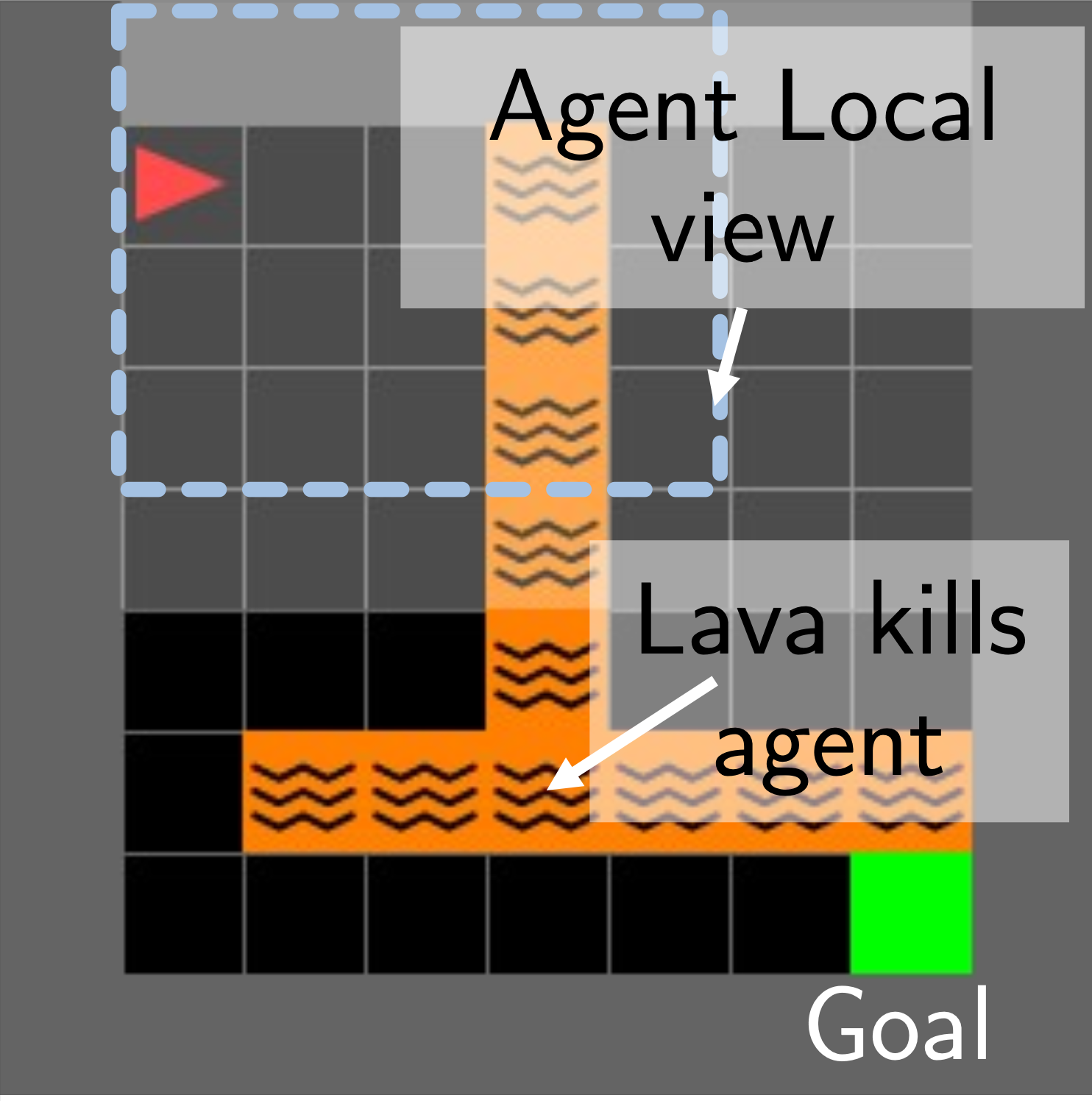}
        \vspace{-11pt}
        \caption{Lava Crossing}
        \label{fig:lava_repr}
    \end{subfigure}
    \hfill
    \begin{subfigure}[b]{0.27\textwidth}
        \includegraphics[width=\linewidth]{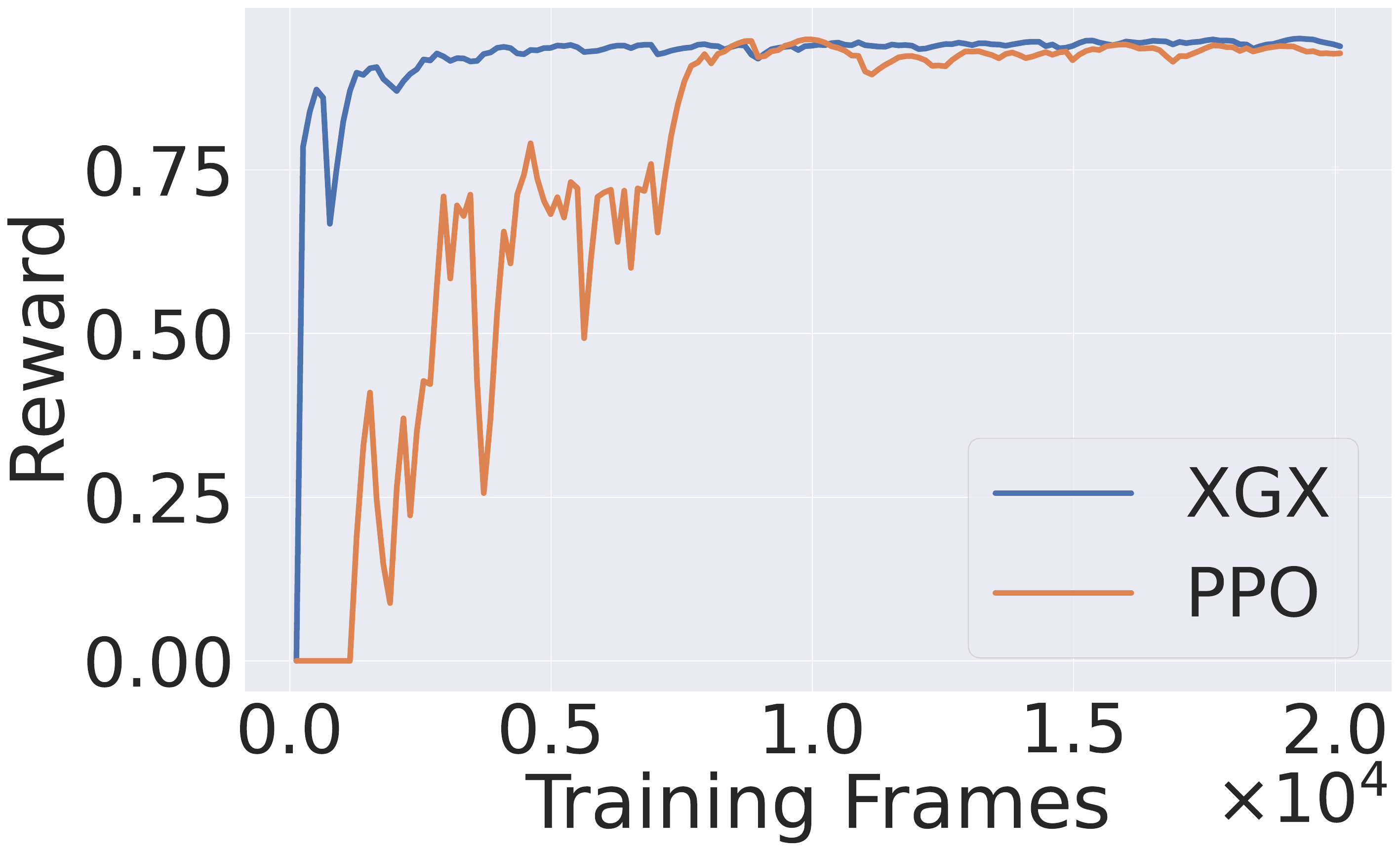}
        \caption{Reward curves on Empty}
        \label{fig:empty}
    \end{subfigure}
    \hfill
    \begin{subfigure}[b]{0.27\textwidth}
        \includegraphics[width=\linewidth]{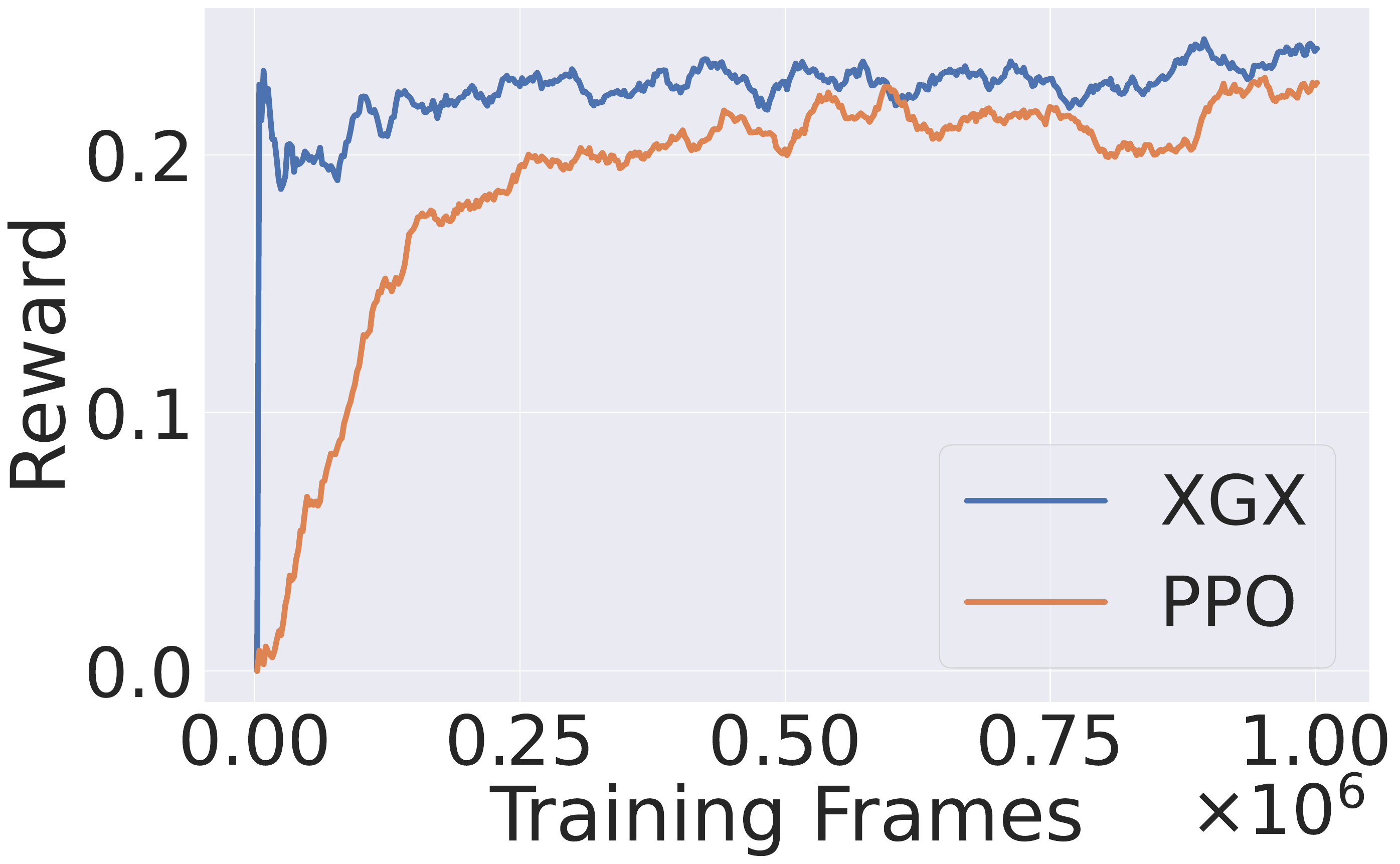}
        \caption{Reward curves on 4-Rooms}
        \label{fig:four}
    \end{subfigure}
    \hfill
    \begin{subfigure}[b]{0.27\textwidth}
        \includegraphics[width=\linewidth]{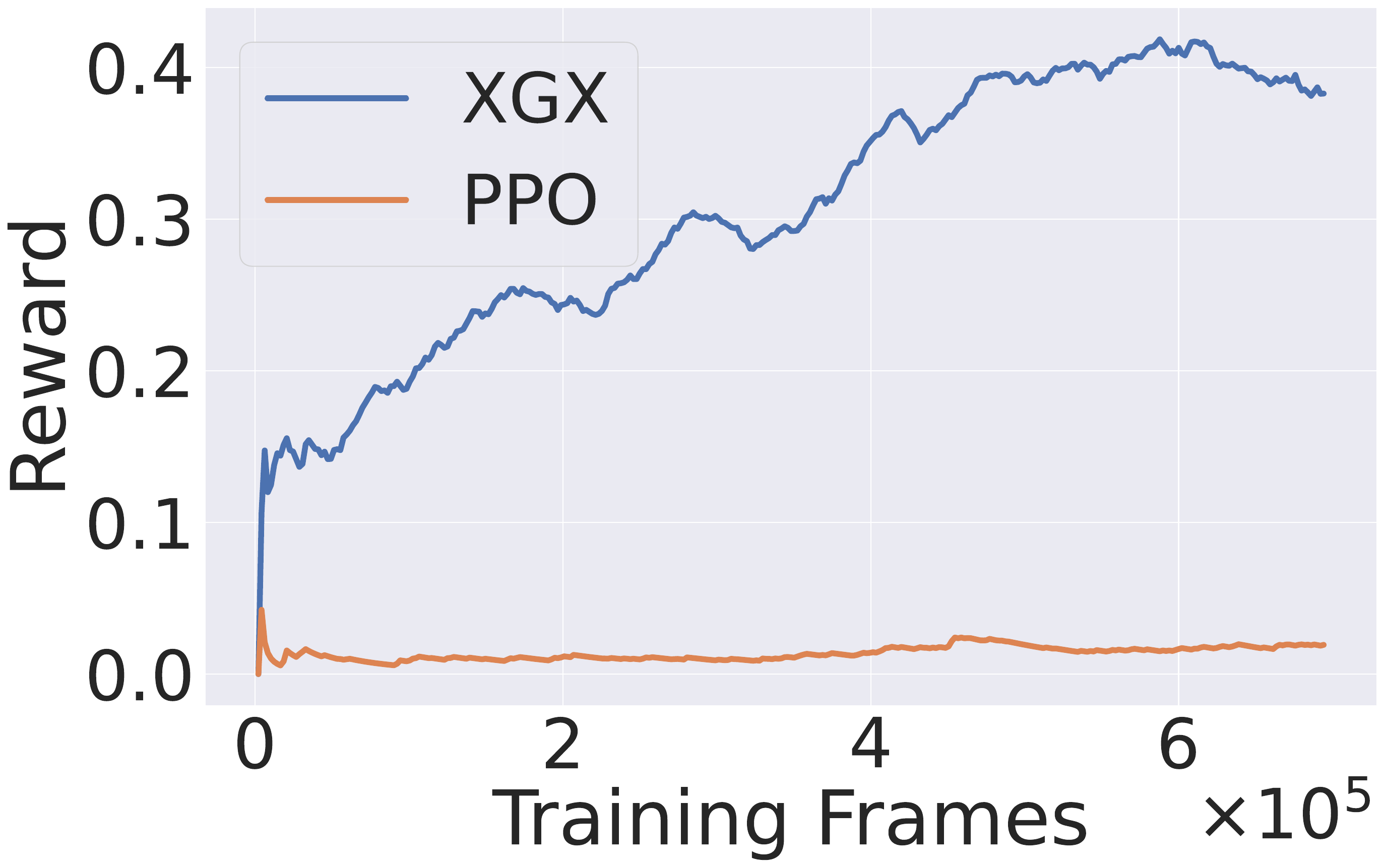}
        \caption{Reward curves on Lava Crossing}
        \label{fig:lava}
    \end{subfigure}
    \vspace{-1mm}
    \caption{(a) one of the three navigation tasks \ourmethodshort is tested on MiniGrid Platform. (b,c,d) Across these three tasks, \ourmethodshort outperforms PPO and trains faster. As task complexity increases (b$\rightarrow$c$\rightarrow$d), the gains from \ourmethodshort become increasly prominent.}
    \vspace{-5mm}
    \label{fig:minigrid_results}
\end{figure*}
\subsection{Exploitation Module Teacher-Forcing (Guidance)}
The high-level objective of a policy-gradient approach like \ourmethodshort with PPO is to learn a policy towards maximizing expected, discounted, cumulative reward. The clipped surrogate loss objective from PPO~\cite{schulman2017proximal} is as follows:
\begin{align}
    \label{eq:xgx}
    \begin{aligned}
    \mathcal{L}(\theta) = \E_{\mathcal{D}}\Bigl[\min(&{p_t}(\theta)\hat{A_t}(\theta), \\
    &\text{clip}\bigl({p_t}(\theta), 1-\epsilon, 1+\epsilon\bigr)\hat{A}_t(\theta)\Bigr],
    \end{aligned}
\end{align}
where $\hat{A_t}(\theta)$ is the advantage function, $\hat{V}_t(\theta)$ is the value estimate from the critic head, $p_t(\theta)$ is ratio function to correct for stale policy used for rollouts, and $\mathtt{clip}$ is performed to keep ratio function within $\epsilon$ dispersion of 1 (prevents too large of a gradient). The  replay buffer of rollouts is denoted by $\mathcal{D}$ consisting of $(s_t,a_t,r_t)$ tuples for state, action, and environment rewards. The modification that allows \pig to guide \piex is rooted in the ratio function. In standard PPO, the ratio function is $p_t(\theta) = \frac{{\pi}(a_t|s_t; \theta)}{{\pi}(a_t|s_t; \theta^{-})}$, where $a_t\sim{\pi}(a_t|s_t; \theta^{-})$ and $\theta^{-}$ is the stale parametrization of policy used to collect rollouts in $\mathcal{D}$. In \ourmethodshort, actions in the replay buffer ($a_t$) are instead sampled from a teacher-forced policy. Concretely:
\begin{align}
    \label{eq:pibar}
    a_t = 
    \begin{cases}
        a_t^{\text{exploit}}\sim\pi_{\text{exploit}}(a|s) & \text{if}\quad s \in \mathcal{S}_{\text{exploit}}\\
        a_t^{\text{explore}}\sim\pi_{\text{explore}}(a|s;\theta^{-}) & \text{otherwise}
    \end{cases},
\end{align}
where \piex, $\mathcal{S}_{\text{exploit}}$, and \pig were introduced in \secref{subsec:explore}, \secref{subsec:phase}, and \secref{subsec:exploit}, respectively. The value function relies on behavior policy as well. Since we employ the above teacher-forced policy, \ourmethodshort value (and thereby advantage) estimates are also different. 
Notably, with \pig employed in \equref{eq:pibar}, the time horizon of the RL formulation shortens significantly. This allows more effective learning from sparse success cues to optimize the parameters $\theta$. Exploitation module \pig is specialized for getting to the goal, it increases the reward the agent receives overall during training. Another advantage of the decomposition in \equref{eq:pibar}, also allows for back-tracking, \ie, if the agent leaves `exploitation states' $\mathcal{S}_{\text{exploit}}$, the agent returns to following the exploration module.

We choose to base \ourmethodshort on PPO~\cite{schulman2017proximal} for two practical reasons. First, it is the de facto algorithm for most RL formulations, with a stable convergence. Second, this choice allows us to build off the excellent support the AIHabitat photorealistic simulator has for distributed and decentralized PPO~\cite{wijmans2019dd}.

\section{Experiments}
\label{sec:experiments}

In this section, we test our \ourmethodshort methodology across (1) three 2D gridworld tasks, (2) large-scale photorealistic simulation across 1000 indoor environments, (3) physical robot runs across three diverse real-world scenes.
\subsection{Diagnostic Task: Navigation in 2D MiniGrid}

We investigate \ourmethodshort in the fast MiniGrid environments~\cite{MinigridMiniworld23}, commonly used for proof of concept in prior works of visual navigation~\cite{WeihsJain2020Bridging,parisi2022unsurprising,jain2021gridtopix,parisi2021interesting}. Particularly, three tasks we benchmark on are: Empty, 4-Rooms, and Lava Crossing. These tasks require the navigation agent to explore the environment, avoid obstacles (particularly the detrimental lava) and stop at the goal represented as a green square. The standard reward structure used in these tasks is $1 - 0.9 * \frac{\text{step count}}{\text{max steps}}$ where max steps is the maximum number of steps the agent is allowed to take in a given episode. The agent observes a local, `egocentric' (along the direction the agent is facing) patch of the 2D maze. An example Lava Crossing environment (mazes are procedurally generated and randomized) with helpful annotations is visualized in~\figref{fig:lava_repr}.
For \ourmethodshort's exploitation module \pig, the agent uses a local planner to create an action that will navigate it toward the goal (only when goal is agent's egocentric view). 

\noindent\textbf{Results.}
As demonstrated in~\figref{fig:minigrid_results}, \ourmethodshort unanimously outperforms PPO across the three tasks. \ourmethodshort demonstrates faster convergence over vanilla PPO. In the empty environment task, \ourmethodshort achieves a reward of $0.9$ after just 5\% of training budget, while PPO needs 40\%. 
In the Lava Crossing environment, \ourmethodshort attains a reward of $0.41$ while PPO is stuck at $\sim0.01$ despite training it for over $500k$ steps.

\subsection{Object-Goal Navigation Task} %
\label{sec:task}
We experiment with \pipe are on semantic visual navigation task of Object-Goal Navigation or \onav~\cite{batra2020objectnav,anderson2018evaluation}. This choice is based on several factors -- reproducible, photorealistic simulation, suite of published baselines, and large-scale suitable for learning-based robotics and sim-to-real transfer. Next, we include a brief task description, information on the associated sensors, and details of the standard metrics.
\begin{table}[t]
\centering
\resizebox{\columnwidth}{!}{%
    \begin{tabular}{l l l c c }
        \toprule
    \# & Method & Type & \success$\uparrow$ & \spl$\uparrow$\\
    \midrule
    1 & DD-PPO~\cite{wijmans2019dd} & RL & 27.9 & 14.2 \\
    2 & OVRL~\cite{yadav2022OVRL} & SSL$\rightarrow$RL  & 62.0 & 26.8\\
    3 & OVRL-V2~\cite{yadav2023ovrl} & SSL$\rightarrow$RL  & 64.7 & 28.1\\
    4 & RRR~\cite{raychaudhuri2023reduce} & Modular TL & 30.0 & 14.0\\
    5 & Frontier Based Expl.~\cite{yenamandra2023homerobot, gervet2022navigating} & Classical & 26.0 & 15.2 \\
    6 & Habitat-Web~\cite{ramrakhya2022habitat} (paper) & IL  &  57.6 & 23.8\\
    7 & Habitat-Web~\cite{ramrakhya2022habitat} (our impl.) & IL  &  64.1 & 27.1 \\
    8 & PIRLNav~\cite{ramrakhya2023pirlnav} (paper) & IL$\rightarrow$RL  & 61.9 & 27.9\\
    9 & PIRLNav~\cite{ramrakhya2023pirlnav} (our impl.) & IL$\rightarrow$RL & 70.4 & 34.1\\
    10 & \ourmethodshort (ours) & IL$\rightarrow$RL  & \textbf{72.9} & \textbf{35.7} \\
    \midrule
    \multicolumn{5}{c}{RL: reinforcement, IL: imitation, SSL: self-supervised, TL: transfer}\\
    \bottomrule
    \end{tabular}
    }
    \caption{\textbf{Quantitative results for semantic navigation on \onav}. 
\ourmethod (\ourmethodshort) significantly outperforms prior works based on IL, RL, and self-supervised representation learning. An equivalent jump in SPL is much harder than the same jump in success rate, indicating a significantly more efficient planning by \ourmethodshort. $\uparrow$ denotes higher is better.}
\label{tab:habitat}
\vspace{-5mm}
\end{table}

\noindent\textbf{Task Definition.} We adopt the protocol laid out for \habitat~\cite{batra2020objectnav} which has been standardized as a public benchmark as well~\cite{habitat-challenge-2021}. At the start of an \onav episode, an embodied agent is initialized at a random location and is tasked to navigate to a goal category. At every time step the agent can choose an action from the action space $\mathcal{A} :=$ \{\textit{move forward, turn right, turn left, stop, look down, look up}\}. The episode is considered a success if the agent can navigate within 1 meter of an instance of the goal category and execute the \textit{stop} action. Consistent with prior work, we too adopt a sparse reward structure that returns a 1 on success and a 0 otherwise, at every time step.
The goal definition, \ie, the category of object to navigate to is sampled from a set of 6 categories~\cite{batra2020objectnav} that are visually and physically well-defined. 
Consistent with several works on \onav~\cite{batra2020objectnav,majumdarzson,yadav2022OVRL,rramrakhya2022,al2022ZER}, we adopt the Habitat-Matterport3D (HM3D) scenes dataset. 
HM3D consists of high-quality photorealistic scans of 1000 real-world indoor scans. 
HM3D is an ideal choice for our study as it includes several scene types -- homes, workspaces, eateries, and retail shops. Scenes span many geographical locations, floor area, and multiple floored scenes are also included. This diversity allows for a better chance at generalizing to real-robot runs.
For evaluation, we adopt the standard and publicly available 2000 episodes HM3D-val split.

\noindent\textbf{Sensors.}
We adopt the standard sensor suite for \habitat \onav~\cite{habitat-challenge-2020,habitat-challenge-2021}. Particularly, the agent has access to three sensor observations: (1) egocentric RGB image of 640$\times$480, (2) corresponding depth mask, (3) relative localization obtained from a `GPS+Compass sensor'~\cite{habitat19iccv,batra2020objectnav}. For off-the-shelf $f_{\text{semantic}}(\cdot)$, we utilize a RedNet~\cite{jiang2018rednet} trained on the HM3D-training split to predict egocentric semantic information in simulation. In the real-world we utilize DETIC~\cite{zhou2022detecting} to predict semantic segmentation.

\noindent\textbf{Metrics.} Following the literature in embodied navigation~\cite{habitat19iccv,wijmans2019dd,jain2019CVPRTBONE,JainWeihs2020CordialSync,chen2019audio,batra2020objectnav,raychaudhuri2021language}, we adopt metrics of success rate (\success) and success weighted by path length (\spl). \spl captures the policy's efficiency in path planning. For targeted evaluation of exploration modules, we also report a metric from learning-based exploration~\cite{chen2018learning,Chaplot2020Explore,patel2021comon,chaplotNeuralTopologicalSLAM2020,ramakrishnan2020exploration}, particularly, the percent of the environment seen (\percentcov). A point in the environment is considered `covered' if it is within $3.2$m of the agent and its field of view. Note, less coverage is better only if success rate is the same. To this end, symmetric to the \spl metric introduced in~\cite{anderson2018evaluation}, we devise a metric (\scov) to balance task success with efficient exploration, defined as:
\begin{align}
    \vspace{-1mm}
    \label{eq:scov}
   \frac{1}{N}\sum_{i=1}^{N}S_i\frac{\%~\mathtt{Cov}_i^{\text{oracle}}}{\text{max}(\%~\mathtt{Cov}_i, \%~\mathtt{Cov}_i^{\text{oracle}})}
   \vspace{-1mm}
\end{align}

Where N is the number of total episodes, $S_i$ specifies the success of episode i, and $\%~\mathtt{Cov}_i^{\text{oracle}}$ is the \percentcov of a given episode when using an shortest-path policy. 

\subsection{Methods}
\label{sec:baselines}
We benchmark utilizing a diverse set of baselines spanning purely reinforcement learning, imitation learning, self-supervised representation learning, transfer learning, classical baselines, and combinations of the former:\\
$\bullet$~DDPPO~\cite{wijmans2019dd}: Purely RL baseline that employs proximal policy optimization~\cite{schulman2017proximal} in distributed and decentralized manner. DDPPO is a widely adopted deep RL baseline in prior works, across task definitions~\cite{ye2021auxiliary,hahn2021no, al2022ZER,yadav2022OVRL,khandelwal2022simple}, and perfectly solved point-goal navigation task~\cite{habitat19iccv}.\\
$\bullet$~OVRL~\cite{yadav2022OVRL}: Powered by self-supervised representation learning, OVRL pretrains a modified ResNet50~\cite{he2016deep} head on the Omnidata Starter Dataset~\cite{eftekhar2021omnidata}. Using this initialization, Yadav~\etal finetune the encoder and the policy on top using 500M frames of experience.\\
$\bullet$~OVRL-V2~\cite{yadav2023ovrl}: Similar to OVRL, but the choice of the encoder is changed to vision transformers (ViT)~\cite{dosovitskiy2020image} with a compression layer through a masked auto-encoding~\cite{he2022masked}. 500M steps in the simulator are used for finetuning\jwextra{ both the ViT and the policy head}.\\ 
$\bullet$~Habitat-Web~\cite{ramrakhya2022habitat}: Purely IL on $\sim$80k human-collected episodes~\jwextra{using a simulation-web interface}. Utilizes an inflection-weight~\cite{wijmans2019embodied} to avoid learning trivial policies.\\
$\bullet$~PIRLNav~\cite{ramrakhya2023pirlnav}: Going a step further from Habitat-Web baseline, this adopts a two-stage training regime -- imitation learning for 500M steps followed by PPO updates for 300M frames. This is shown to improve performance in simulation.\\
$\bullet$~RRR~\cite{raychaudhuri2023reduce,raychaudhuri2023mopa}: Modular policy with a zero-shot transfer learning from an HM3D point-goal module to an object-goal navigation agent.\\
$\bullet$~FBE~\cite{yenamandra2023homerobot, gervet2022navigating}: Frontier-Based exploration while incorporating semantics to find the goal; adopted by Gervet~\etal~\cite{gervet2022navigating}. We report results on full HM3D validation split (not just environments with a single floor~\cite{gervet2022navigating}).

\begin{table}[t]
\centering
\vspace{3mm}
\resizebox{\columnwidth}{!}{%
    \begin{tabular}{l l c c}
    \toprule
    \# & Method & \success$\uparrow$ & \spl$\uparrow$ \\
    \midrule
    1 & Init. with IL on Human Demos + Fine-Tune~\cite{ramrakhya2023pirlnav} & 70.4 & 34.1 \\
    2 & 1 + Policy Decomposition & 71.5 & 34.3\\
    3 & 1 + Policy Decomposition + Guidance (\ie \ourmethodshort) & \textbf{72.9} & \textbf{35.7}\\
    \midrule
    {\color{gray}4} & {\color{gray} \ourmethodshort + GT Semantics \textit{(upper bound)}} & {\color{gray}73.5} & {\color{gray}39.8} \\
    \bottomrule
    \vspace{-5mm}
    \end{tabular}
    }
    \caption{\textbf{Head-on ablations for \ourmethodshort }. We observe steady gains in performance by including the exploitation module \pig in our policy decomposition and adding \textit{guidance} to neural exploration module coming from teacher-forced policy optimization.}
\label{tab:ablations}
\end{table}

\subsection{Results and Analysis (Photorealistic Simulation)}
\label{subsec:experiments_simulation}
Metrics from empirical runs in \habitat are reported in \tabref{tab:habitat}, \tabref{tab:ablations}, and \tabref{tab:exploration}. We discuss these results and analysis in the following text. As detailed in~\secref{sec:task}, all results are reported on the HM3D validation scenes (never seen during training). 

\noindent\textbf{\ourmethodshort improves over all prior IL, RL, and SSL baselines~(\tabref{tab:habitat}).} Compared with diverse baselines (\secref{sec:baselines}), \ourmethodshort (in row 10) demonstrates significant gains, improving $70.4\rightarrow72.9$ in success rate (relative $4\%\uparrow$) over the previous best IL+RL method -- PIRLNav~\cite{ramrakhya2023pirlnav} (in row 9). As we intuitively guessed, communicative exploration-exploitation via \ourmethodshort has more efficient path planning, by reducing the `learning load' on the exploration module. This helps improve the challenging metric of SPL by a relative $5\%$ ($34.1\rightarrow35.7$). The best SSL method, OVRL-V2~\cite{yadav2023ovrl} (in row 3), is fairly competitive in success rate ($64.7$ \vs $72.9$) but significantly lags behind on SPL ($28.1$ \vs $35.7$). Notably, \ourmethodshort also outperforms classical frontier-based methods (row 5), improving from 26.0\% to our 72.9\% success rate.

\noindent\textbf{Both the exploitation module and guidance are helpful~(\tabref{tab:ablations}).} Two key aspects of our proposed approach are (1) exploitation module: based on basic visuo-motor servoing and geometric computer vision and (2) guidance from this module to exploration. Here we undertake head-on ablations to quantitatively evaluate the utility of both in \tabref{tab:ablations}. We observe improvements of $70.4\rightarrow71.5\rightarrow72.9\%$ (row $1\rightarrow2\rightarrow3$) in success rate by successively adding the policy decomposition and guidance to the initialization. These trends are even more pronounced for the efficiency metric of SPL. This is intuitive as geometric vision-based servoing is almost perfect in path planning when fired correctly. We observed the same trend when adding \ourmethodshort to other models like Habitat-Web~\cite{ramrakhya2022habitat} (seconding our results with PIRLNav~\cite{ramrakhya2023pirlnav} in \tabref{tab:ablations}.
\jwextra{If accurate semantics are available, \ourmethodshort can scale with them to improve performance to 73.5\% (row 4). This shows \ourmethodshort can successfully tap into improvements in vision towards better navigation systems.}

\noindent\textbf{Efficiency Analysis -- \ourmethodshort can explore faster~(\tabref{tab:exploration}).}
\jwextra{Here we undertake a focused investigation of learned-based exploration i.e. quantifying exploration and not navigation success to the goal.}
We find that \ourmethodshort explores an average of $59.8\%$ per scene (compared to PIRLNav's $63.5\%$), while still performing better on other navigation success metrics. 
This is because \ourmethodshort's exploration module was \textit{a priori} optimized to offload servoing to the goal of the exploitation module. \ourmethodshort taps into compositionality, making exploration significantly more efficient while also improving performance.
This intuition is also empirically supported by \ourmethodshort outperforming PIRLNav on \scov at 36.3\% to 39.1\%.

\begin{table}[t]
\centering
\vspace{2mm}
    \begin{tabular}{l  c c }
    \toprule
    Method & \scov $\uparrow$  & \cov  \\
    \midrule
    PIRLNav~\cite{ramrakhya2023pirlnav} & 36.3 & 63.5\\
    \ourmethodshort (ours) & \textbf{39.1}  & 59.8\\
    \bottomrule
    \end{tabular}
    \caption{\textbf{Analysis for exploration efficiency.} We find that \ourmethodshort outperforms PIRLNav (row 9 and row 10, \tabref{tab:habitat}) and it does so with more efficient, goal-conditioned exploration.}
    \vspace{-3mm}
\label{tab:exploration}
\end{table}

\begin{figure}[t]
    \centering
    \includegraphics[width=0.48\textwidth]{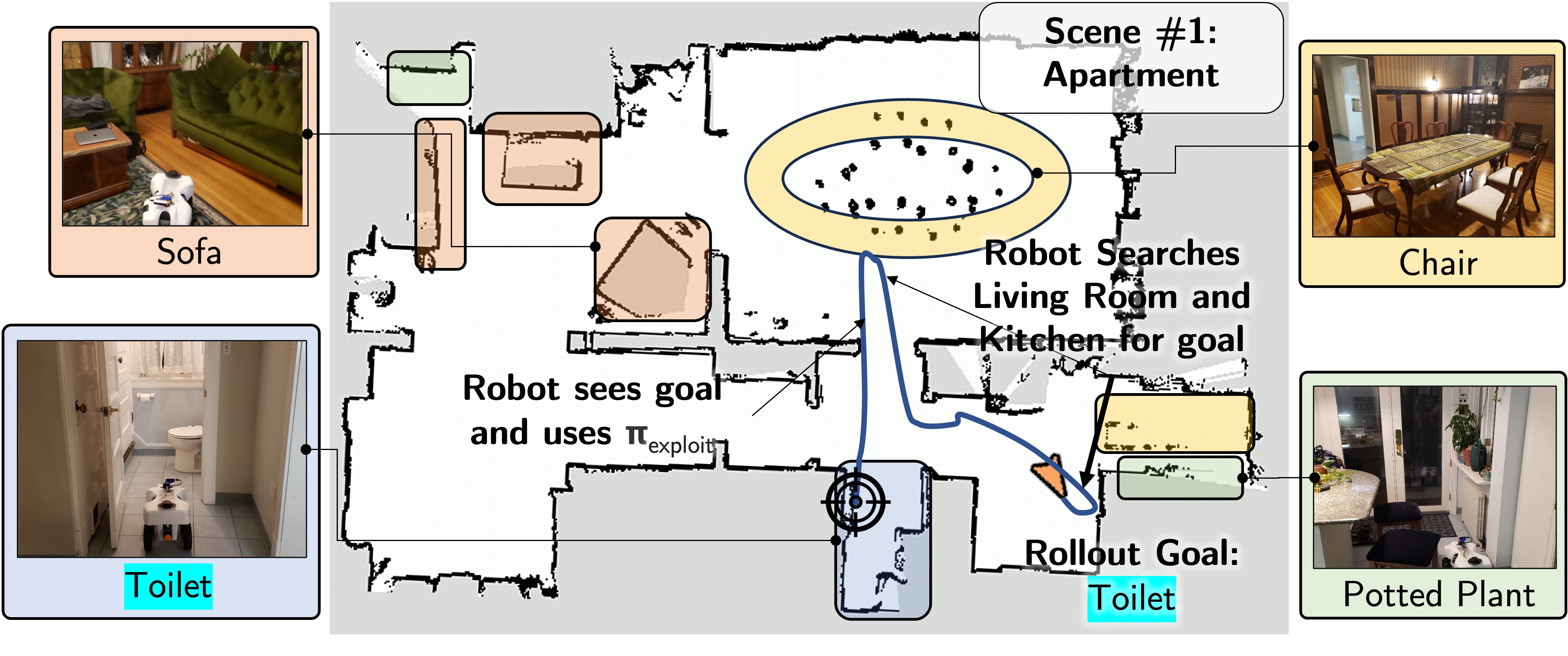}
    \caption{\textbf{Physical robot navigation setup and rollout visualization in $\mathtt{Apartment}$  environment.} The agent navigates to one of 4 goal categories. Here, we visualize an actual rollout for the `toilet' category and mark the phase transition to the exploit module. Note: The agent has not seen the scene or access the top-down map (the map is for visualization).}
    \vspace{-3mm}
    \label{fig:experiments_robot_vis}
\end{figure}

\begin{figure}[t]
    \centering
    \vspace{2mm}
    \includegraphics[width=0.96\linewidth]{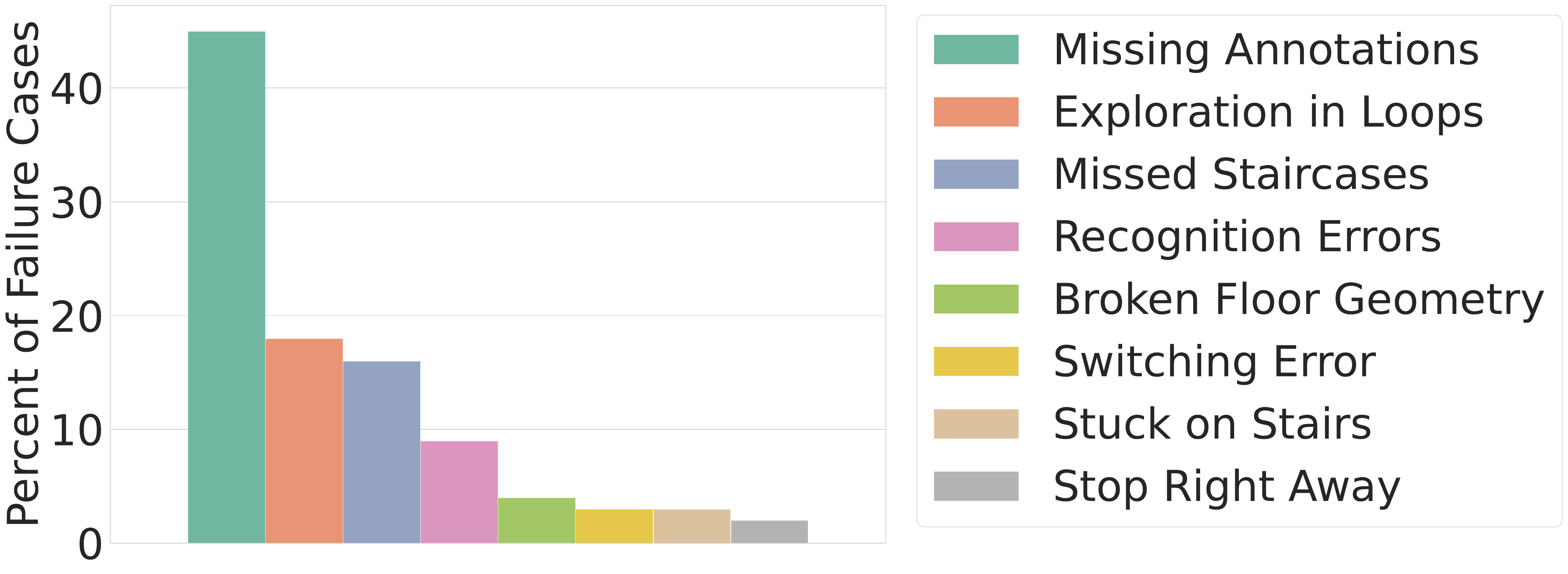}
    \caption{\textbf{Error Analysis of  \ourmethodshort \textit{(simulation)}.} Beyond missing annotations in the dataset, the top three error modes of \ourmethodshort are (1) \piex manifesting a looping behavior, (2) missing staircases, (3) and errors in semantic segmentation.}
    \label{fig:xgxfail}

\end{figure}

\noindent\textbf{Error Analysis.}
The failure modes of \ourmethodshort are visualized in~\figref{fig:xgxfail}. The largest source of failure is missing annotations, where $f_{\text{semantic}}$ of RedNet would correctly classify a goal object, only for the environment's semantic mesh to not be correctly labeled for the given object.  Major failure modes for the exploration module are due to not searching the environment well by either just looping over the same space (`Exploration in Loops') or not trying to go up/down stairs when it should (`Missed Staircases'). Furthermore, the exploration module will occasionally call the `stop' action immediately at the start of an episode (`Stop Right Away'). Issues with the switching mechanism includes failing to correctly recognize the goal (`Recognition Errors') and failing to switch between the exploration and exploitation modules (`Switching Error'). Other simulator errors include the floor geometry not being connected between spaces (`Broken Floor Geometry') and incomplete meshes where if the agent tries to go up/down stairs it can not (`Stuck on Stairs'). Further discussion of these error modes are given in Appendix~\ref{sec:app-discussion}.

\subsection{Results and Analysis (Physical Robot)}
\label{subsec:robotexps}
For comprehensive details regarding implementation, analysis, and results, please see the supplementary video and our project page \href{https://xgxvisnav.github.io/}{xgxvisnav.github.io} and in Appendix~\ref{sec:app-robot}.
 We sim-to-real transfer PIRLNav and our \ourmethodshort on a wheeled navigation robot employed by navigation works~\cite{higuti2019under,gasparino2022wayfast,ji2022proactive,wasserman2022lastmile}. We navigate to five goal categories \{couch, TV, chair, toilet, potted plant\} across 14 trajectories per method. \jwextra{Following success in robot navigation and locomotion research~\cite{truong2021learning,kim2019highly,hirose2019deep}, we realize the low-level actuation using Model Predictive Control. }We test our robotics setup in three diverse scenes (1) $\mathtt{Apartment}$ - as visualized in~\figref{fig:experiments_robot_vis} (2) $\mathtt{Office}$ - environment with long hallways and many connecting rooms and (3) $\mathtt{Food~Court}$ - containing many chairs and furniture that needs to be avoided. Visualizations of these environments are given in Appendix~\ref{sec:app-qual}. In order to acquire pose information, we employ localization using the robot's LiDAR sensor and SLAM~\cite{KohlbrecherMeyerStrykKlingaufFlexibleSlamSystem2011}.

\noindent\textbf{Accurate embodiment reduces the sim-to-real gap (\tabref{tab:robot}, rows  1 \& 2).}
The best-performing prior method, PIRLNav, does not demonstrate intuitive behaviors on the robot (see row 1, \tabref{tab:robot}). This is despite us trying several lighting, goal categories, and distance to goal choices. 
We adapt PIRLNav (see row 2) by retraining it with a robot-specific configuration (height and field-of-view) as the default simulator configuration is different. This improved navigation behavior and a successful trajectory rollout. This is consistent with a prior study on end-to-end methods~\cite{gervet2022navigating}. %

\noindent\textbf{\ourmethodshort improves the performance of the robot as well (\tabref{tab:robot}, row 3).}
In \tabref{tab:robot}, we find that \ourmethodshort empirically has the best performance in physical robot experiments as well. Comparing rows 2 and 3, we observe an improvement from $8.4\% \rightarrow 23.3\%$ in SPL.

\begin{table}[h]
\centering
\vspace{2mm}
\begin{tabular}{llcc}
    \toprule
    &Method & \success$\uparrow$ & \spl$\uparrow$\\
    \midrule
    1&PIRLNav (released ckpt) & $0.0$ & $0.0$\\
    2&PIRLNav w/ our adaptation & 14.2 & 8.4\\
    3&\ourmethodshort (ours) & \textbf{35.7} & \textbf{23.3}\\
\bottomrule
\end{tabular}
\caption{\textbf{Real world object-goal navigation results.} In real world robotics experiments, \ourmethodshort performs the best.} %
\vspace{-4mm}
\label{tab:robot}
\end{table}

\section{Conclusion}
\label{sec:conclusion}
In this work, we ask a fundamental question: ``Can decomposed navigation agents learn better exploration when guided by their exploitation abilities?'' To this end, we introduce \ourmethodshort which utilizes teacher forcing from the exploitation to the exploration module, implemented via off-policy updates. We deploy \ourmethodshort with standard choices of parametric exploration and simple geometric exploitation modules, leading to state-of-the-art performance in simulation, quantifiable efficiency gains in exploration, and an over 2-fold improvement in physical robotics experiments.

{ 
  \bibliographystyle{ieee_fullname.bst}
  \bibliography{references}  }

\clearpage

\section*{Appendix -- Exploitation-Guided Exploration for Semantic Embodied Navigation}
\begin{itemize}\compresslist
    \item [\ref{sec:app-robot}] More details regarding the hardware and software used in our real-world robotics experiments. Visualizations of the real-world experiments are also included.
    \item [\ref{sec:app-qual}] Further real-world qualitative robot figures and rollout. Also discusses the use of robot parameters in simulation.
    \item [\ref{sec:app-impl}] Extra details for implementing \ourmethodshort, such as hyperparameter choice and some details of the low level policy.
    \item [\ref{sec:app-discussion}] Further details of the failure modes of the error modes of \ourmethodshort.
\end{itemize}
\appendix

\subsection{Robot Experiments Implementation Details  (Supplements \secref{subsec:robotexps})}
\label{sec:app-robot}

\noindent\textbf{Hardware details.} Key features in the wheeled robot: (1) four independent trains, (2) an upward facing camera, and (3) zero radius turns help us in the policy transfer. 
We utilize an economical, widely-used RGB camera (Intel\textsuperscript{\textregistered} RealSense\textsuperscript{\texttrademark} D435i), a 2D LIDAR (Hokuyo\textsuperscript{\textregistered} UST-10LX\textsuperscript{\texttrademark}), that in conjunction with Hector SLAM~\cite{KohlbrecherMeyerStrykKlingaufFlexibleSlamSystem2011} helps localize the agent in the real-world (analogous to the `GPS+Compass sensor' in \habitat). At the end of every episode, the map is deleted. Goal categories are \{chair, couch, tv, toilet, potted plant\} (overlapping with HM3D~\cite{ramakrishnan2021hm3d} objects in \habitat).
Following success in robot navigation and locomotion research~\cite{truong2021learning,kim2019highly,hirose2019deep}, we realize the low-level actuation using Model Predictive Control. 

\noindent\textbf{Adapting embodiment for bridging sim-to-real gap.} In row 2 and 3 of \tabref{tab:robot}, we train the agent in the same \habitat scenes but with new height, size, and camera parameters from the robot hardware. 
In order to train the model to the new physical parameters, we initialize the network weights with the released PIRLNav checkpoint and perform PPO updates for 600M frames. This simple adaptation of network weights leads to gains in performance, as we reported in~\tabref{tab:robot}.

\begin{figure}[t]
    \centering
    \includegraphics[width=0.9\linewidth]{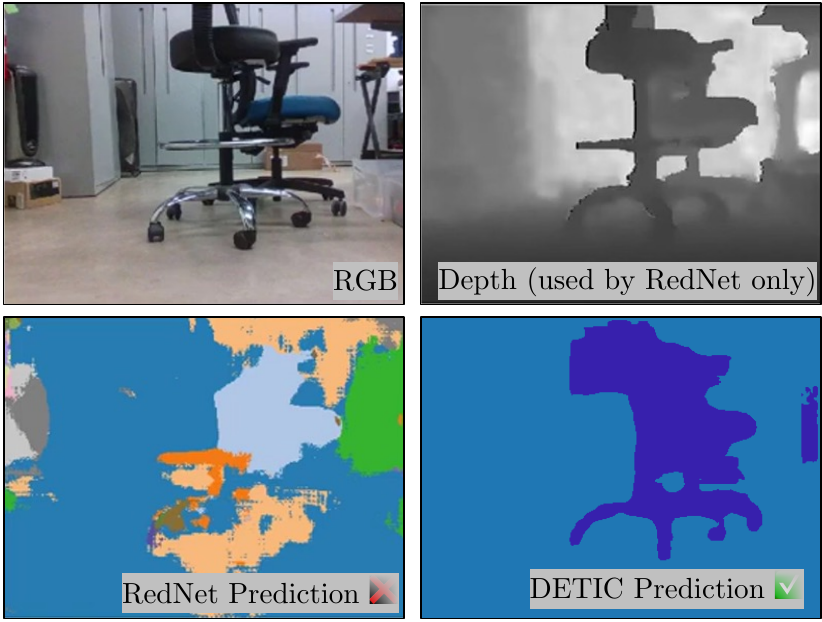}
    \caption{\textbf{We find choice of detector is crucial for sim-to-real generalization ($f_{semantic}$ in \secref{sec:approach}).} 
    Here, we deploy RedNet~\cite{jiang2018rednet} and Detic~\cite{zhou2022detecting} from real-world images collected on the robot. 
    RedNet is \textit{brittle} to being trained in simulation and perhaps expects smooth depth information. Not reliant on depth, we found Detic to be \textit{robust} across hours of robot learning experiments.}
    \label{fig:robotdetect}
    \vspace{-4mm}
\end{figure}
\begin{figure*}[t]
    \centering
    \includegraphics[width=0.99\textwidth]{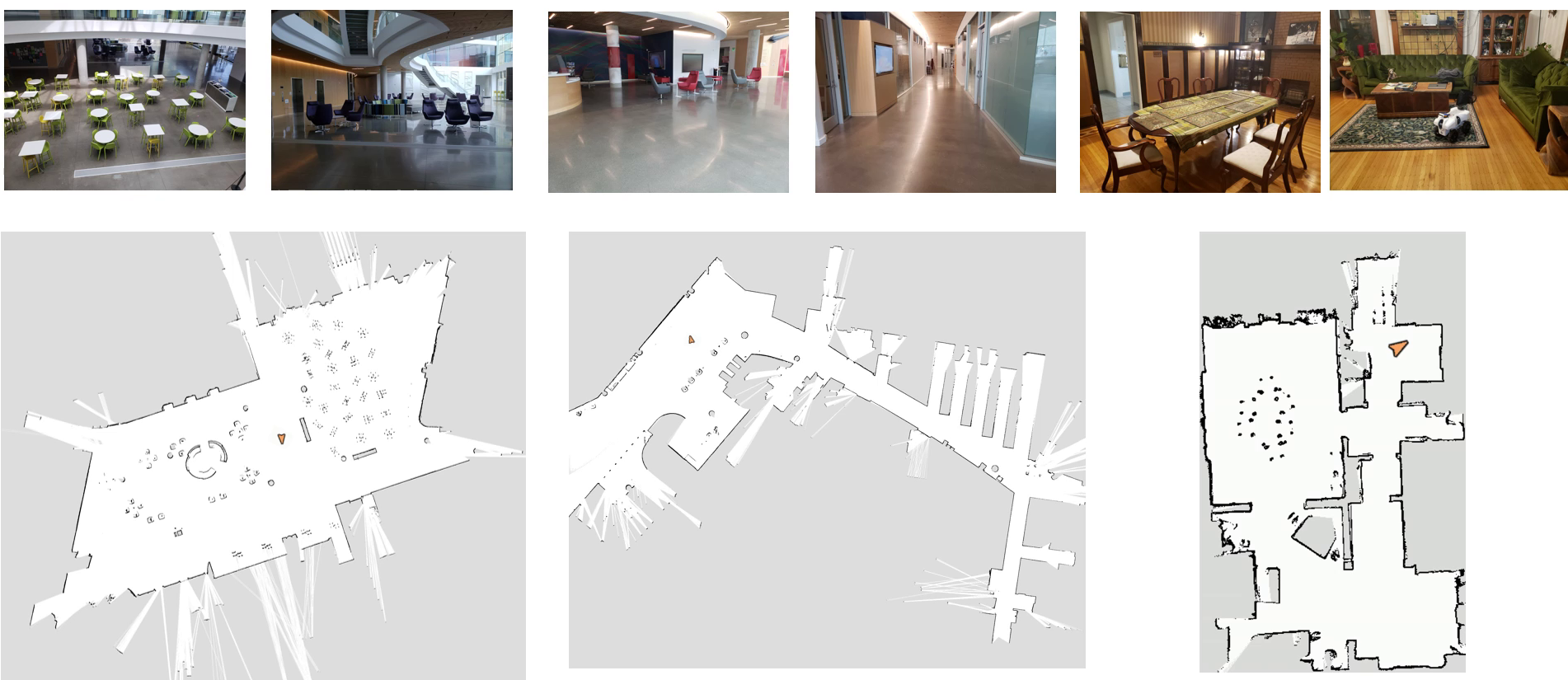}
    \caption{\textbf{Real-world Visualization.}  Visualizations of the $\mathtt{Food~Court}$, $\mathtt{Office}$, and $\mathtt{Apartment}$ environments that the robot is tested on.}
    \vspace{-3mm}
    \label{fig:realworldvis}
\end{figure*}

\begin{figure*}[t]
    \centering
    \includegraphics[width=0.99\textwidth]{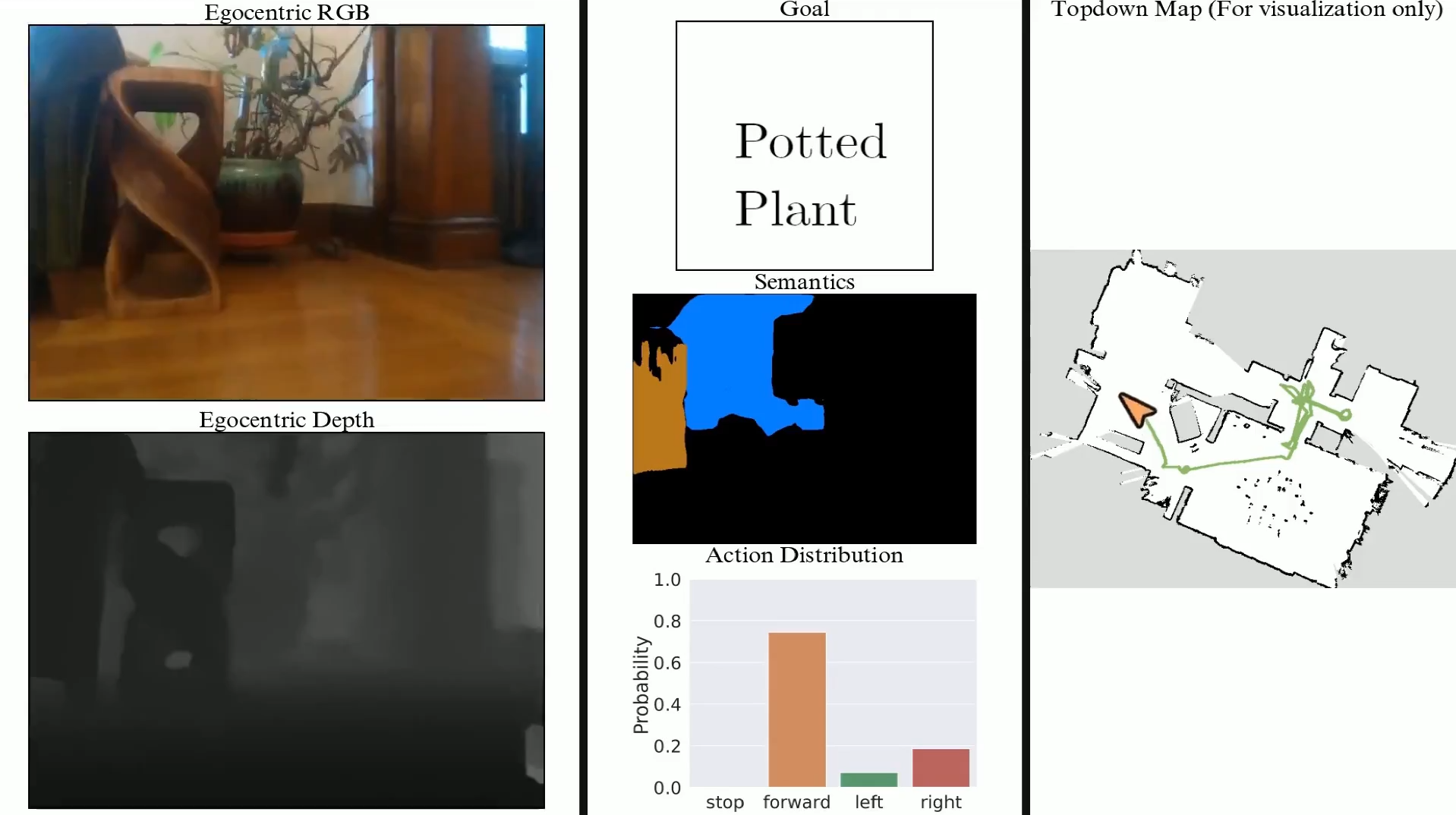}
    \caption{\textbf{Real-world Rollout Example.} The robot is tasked with reaching the potted plant goal. The robot moves through multiple rooms before eventually reaching the goal. The egocentric RGB and depth are visualized on the left hand side. The middle column contains the goal, the output of the semantic prediction from DETIC, and the previously predicted action distribution from the exploration policy. The right column contains a visualization of the top-down map, with a green trajectory marking where the robot traveled from during the rollout.}
    \vspace{-3mm}
    \label{fig:rolllout}
\end{figure*}

\subsection{Qualitative Robot Results (Supplements \secref{subsec:robotexps})} 
\label{sec:app-qual}
\noindent\textbf{Real-world environment visualizations.} In \figref{fig:realworldvis} we provide top-down maps and pictures collected from the $\mathtt{Food~Court}$, $\mathtt{Office}$, and $\mathtt{Apartment}$ environments. The top-down maps also include the agent in them to provide scale for the environments. In \figref{fig:rolllout} an example of a rollout is provided where the agent navigates to a potted plant goal that is multiple rooms away from the starting location.

\begin{figure}[h]
    \centering
  \begin{subfigure}[b]{0.225\textwidth}
    \centering
    \includegraphics[width=\linewidth]{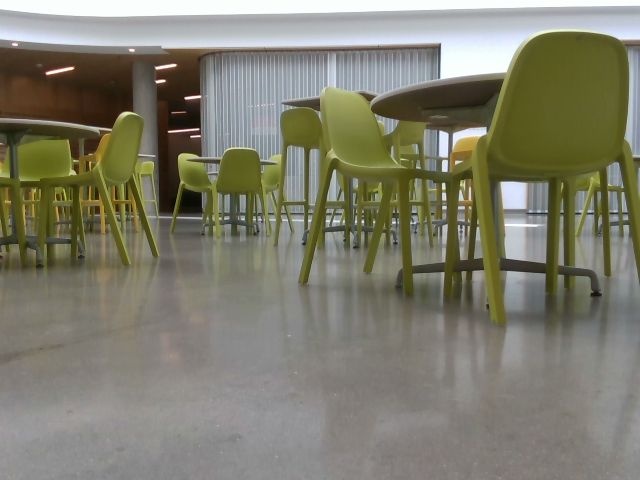}
    \caption{PIRLNav with simulator parameters fails to stop at a chair.}
    \label{fig:supp_qual_robot_pirl}
    \end{subfigure}
    \hspace{0.03\linewidth}
    \begin{subfigure}[b]{0.225\textwidth}
    \centering
    \includegraphics[width=\linewidth]{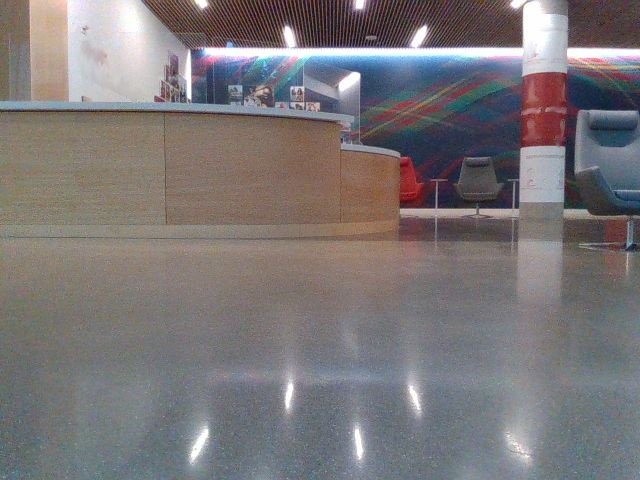}
    \caption{PIRLNav with robot parameters succeeds at arriving at the TV goal.}
    \label{fig:supp_qual_robot_param}
  \end{subfigure}
    \caption{(a) In this setting, our baseline of PIRLNav (released ckpt), trained with \habitat camera and agent parameters fails to stop at a chair. (b) In this scenario, using PIRLNav trained with the robot parameters is able to search the environment for a TV.}
\end{figure}

\noindent\textbf{Different agent parameters in real and simulation causes failure.}
As reported in~\secref{subsec:robotexps}, the baseline implementation of utilizing PIRLNav with Habitat robot and camera parameters failed to ever reach the goal. On our website we show a video of this method on the scenario given in~\figref{fig:supp_qual_robot_pirl} failing to reach ANY of the chairs. This is due to the policy never calling the stop action. We attribute this failure not only due to the sim-to-real gap, but also because of the large discrepancy between the embodiment of the agent through its different heights and camera parameters. Retraining this model with the robot's parameters allows our new model to reach a TV, with the starting location shown in~\figref{fig:supp_qual_robot_param}.

\subsection{Policy Implementation Details (Supplements \secref{subsec:exploit})}
\label{sec:app-impl}
\noindent\textbf{Hyperparameters of $\pi_{explore}$.} A full table of our hyperparameters are given in~\tabref{tab:app-hyperparams} which are used by default. For training our policy with PPO we take advantage of Generalized Advantage Estimation~\cite{schulman2015high}. We train using a distributed version of PPO across 64 GPUs, each with 8 environments per GPU. We perform 2 rollouts per batch. 

\noindent\textbf{Waypoint Creation in $\pi_{geom}$.} The exploitation module $\pi_{geom}$ takes advantage of semantics from a semantic segmentation model, \realseg~\cite{zhou2022detecting}. This model was chosen as it has a large potential for sim-to-real transfer over models that are trained specifically for simulation~\cite{jiang2018rednet, cartillier2021semantic}. As shown in~\figref{fig:robotdetect}, the sim-to-real gap is glaring and RedNet's dependence on perfect depth could be a rationale behind it's poor performance. After this, the semantic segmentation mask is applied to the depth map, only keeping depth values corresponding to the goal. A depth value is extracted from the depth values to find a waypoint in 3D. Rather than averaging over the depth values in the mask, or other heuristics, we found that the best performance occured when smallest depth value, $d_\text{min}$, in the depth image was utilized for the creation of the waypoint. More concretely, given a the $(x,y)$ position of $d_\text{min}$ as well as the value of $d\text{min}$ and the agent's camera parameters (focal lengths $f_x,f_y$ and principal points $u, v$), a 2D waypoint ($w_\text{2D}$) to the goal can be calculated as follows. 
\begin{align}
    \label{eq:wp}
    w_\text{2D} = (w_x, w_y) = (\frac{x-u\times d_\text{min}}{f_x}, \frac{y-v\times d_\text{min}}{f_y})
\end{align}

The 2D waypoint is then given to our local policy to predict actions that will drive the agent towards the goal. A new waypoint is recalculated at every time step in order for \ourmethod to be robust to depth noise.
\begin{table}[t]
    \centering
    \resizebox{0.45\textwidth}{!}{%
    \setlength{\tabcolsep}{10pt}
    \begin{tabular}{ll}
        \toprule
        \textbf{Hyperparamter} &  \textbf{Value}\\
        \hline\multicolumn{2}{c}{\textit{PPO}}\\\hline
        Clip parameter ($\epsilon$)~\cite{schulman2017proximal} & $0.2$\\
        Entropy Coefficient & $0.0001$\\
        PPO clip & $0.2$\\
        GAE $\gamma$ & $0.99$\\
        GAE $\tau$ & $0.95$\\
        Value loss coefficient & $0.5$\\
        DD-PPO sync fraction & $0.6$\\
        Optimizer & Adam\\
        Weight Decay & $0.0$\\
        Learning Rate & $1.5 \mathrm{e}{-5}$\\
        \hline
        \multicolumn{2}{c}{\textit{Exploration Policy}}\\\hline
        RNN Type & GRU\\
        RNN Layers & $2$\\
        RNN Hidden Size & $2048$\\
        CNN Backbone & ResNet50\\

        \hline\multicolumn{2}{c}{\textit{Exploitation Module}}\\\hline
        Distance Threshold ($\delta$) & $2.5m$\\
        Max Angle for Forward Action & $50^{\circ}$\\
        \hline\multicolumn{2}{c}{\textit{Training}}\\\hline
        Number of GPUs & $64$\\
        Number of Environments per GPU & $8$\\
        Rollout Length & $64$\\
        Rollouts per Batch & $2$\\
        \bottomrule
    \end{tabular}
    }
    \caption{Structural and training hyperparameters for reproducibility for our visual navigation policy.} 
    \label{tab:app-hyperparams}
\end{table}

\subsection{Limitations and Future Work (supplements \secref{subsec:experiments_simulation}}
\label{sec:app-discussion}
\noindent\textbf{Failure Modes.}

In order to better understand the failure modes of \ourmethodshort, we collect and analyze the failure modes of 100 unsuccessful episodes in Habitat. We list these errors modes below.

\textbf{Missing Annotations} ($45\%$) - Agent navigates to the correct goal object category but the episode is still counted as a failure. This is due to missing annotations in the dataset/simulator. While this may be surprising to the reader, kindly note that PIRLNav [21] also inferred this in their analysis accounting for ~$30\%$ of their errors. We are working with the dataset/simulator team towards a resolution for these.

\textbf{Exploration in loops} ($18\%$): Our learned exploration module does not fully explore the environment, instead it keeps revisiting the explored states of the environment. 

\textbf{Missed staircases} ($16\%$): Some episodes in ObjectNav are multi-floored i.e. the agent may need to take stairs to find the goal. We find the agent not utilizing staircases is another failure mode, mostly due to under-exploration of the environment.

\textbf{Recognition errors} ($9\%$): Instances where the goal object isn’t recognized by the semantic segmentation module.

\textbf{Cannot avoid broken floor geometry} ($4\%$): In some scenes floor reconstruction is not complete. This results in the exploitation module being triggered and repeatedly trying to reach the goal without being able to do so (due to a broken floor).

\textbf{Stuck on stairs} ($3\%$): In some scenes, the stairs are cosmetic, not leading to another floor. However, the agent oblivious of this stay stuck trying to ‘continue on the stairs’.

\textbf{Switching Error} ($3\%$): Switch from exploration to exploitation-guidance doesn’t initiate as the exploring agent doesn’t get close enough to the goal for exploitation to kick in.

\textbf{Stop right away} ($2\%$): Due to the soft policy i.e. sampling from distribution rather than an argmax, a very low probability of calling stop still exists which triggers this error mode.

\noindent\textbf{Implementation Failure Modes}
We include an upfront list of failure modes, limitations, and ways to address these in the future: (1) For PIRLNav on the robot, we observe that a common failure mode is not calling the \textit{stop} action despite reaching close to the goal; (2) In the future, experimenting with a tethered policy head (one for actions and one for stop) could be helpful for `calling stop' failure mode, as indicated in the simulation study by \cite{ye2021auxiliary} ; (3) 
\ourmethodshort employs the \href{https://github.com/facebookresearch/Detic/blob/main/configs/Detic_LCOCOI21k_CLIP_R18_640b32_4x_ft4x_max-size.yaml}{real-time Detic implementation} which compromises on accuracy for better inference speed (which we needed for faster experimentation).
(4) Naively choosing the $d_\text{min}$ to be the smallest depth along the semantic mask worked well in simulation, but caused the majority of failure cases of \ourmethodshort on the robotics platform.
(5) We adapt embodiment in \tabref{tab:robot} by lowering agent's height to that of the robot. Since the Matterport cameras utilized to scan HM3D scenes were higher, the observations might have reconstruction side-effects that might contribute to the lower performance.

\end{document}